%% file: main.tex
\documentclass[conference]{IEEEtran}

\usepackage{graphicx}
\usepackage{amsmath}
\usepackage{subfigure}
\usepackage{hyperref}
\usepackage{bbm}
\usepackage{color}

\begin{document}

\title{Machine Learning Based Prediction and Classification of Computational Jobs in Cloud Computing Centers}
\author{\IEEEauthorblockN{Zheqi Zhu\IEEEauthorrefmark{1},
Pingyi Fan\IEEEauthorrefmark{1}, $Senior Member$, \textsl{IEEE},
}
\IEEEauthorblockA{\IEEEauthorrefmark{1}
State Key Laboratory on Microwave and Digital Communications\\
Tsinghua National Laboratory for Information Science and Technology (TNList),\\
Department of Electronic Engineering, Tsinghua University, Beijing,
P.R. China\\
Emails: zhuzq18@mails.tsinghua.edu.cn, fpy@tsinghua.edu.cn
}
}

\maketitle

\input{0abstract.tex}

\input{0keywords.tex}
\input{1introduction.tex}

\input{3lstm.tex}
\input{4clustering.tex}
\input{5conclusion.tex}
\section{Acknowledgement}

\input{refs.bbl}
This work was supported by National Natural Science Foundation of China (NSFC) NO. 61771283 and the China Major State Basic Research Development Program (973 Program) No.2012CB316100(2).


\end{document}

%% file: 0abstract.tex
\begin{abstract}
With the rapid growth of the data volume and the fast increasing of the computational model complexity in the scenario of cloud computing, it becomes an important topic that how to handle users' requests by scheduling computational jobs and assigning the resources in data center.

In order to have a better perception of the computing jobs and their requests of resources, we analyze its characteristics and focus on the prediction and classification of the computing jobs with some machine learning approaches. Specifically, we apply LSTM neural network to predict the arrival of the jobs and the aggregated requests for computing resources. Then we evaluate it on Google Cluster dataset and it shows that the accuracy has been improved compared to the current existing  methods. Additionally, to have a better understanding of the computing jobs, we use an unsupervised hierarchical clustering algorithm, BIRCH, to make classification and get some interpretability of our results in the computing centers.
\end{abstract}

%% file: 0keywords.tex
\begin{IEEEkeywords}
 data center, cloud computing, LSTM neural network, prediction, clustering.
\end{IEEEkeywords}

%% file: 1introduction.tex
\section{Introduction}

In recent years, the volume of data, the complexity of computational models, and the demand for computing resources are rapidly increasing, which has greatly spawned cloud computing.~\cite{zhang2010cloud}\cite{barbarossa2014communicating} made a detail review of challenges in it. In general, two distinct respects should be considered for clouding computing. The first one is resource driven clouding computing, where it mainly consists of the data sensing process and the information transmission process. For the data sensing process, many kinds of sensors are employed to acquire data from its monitoring environments~\cite{zheng2018swipt}\cite{zheng2019fog}. For information transmission process, it mainly consider how to find proper routings \cite{yao2003neighbor} and some efficient ways~\cite{ma2005queuing}\cite{zhang2008network} to speed up the transmission rates or ad hoc network throughputs from the sensors to the cloud computing centers.  The second one is the request oriented clouding computing, where the users will send their requests to the cloud computing centers and get the feedback from the cloud computing center. It mainly consists of the request transmission process, the job computing process in cloud computing center and the information feedback process.  In this paper, we are mainly focusing on the request-oriented cloud computing.  In fact, cloud computing are usually carried out in computing clusters so that the requests of users can be processed in parallel and the computing efficiency will be higher. That is to say, proper computing clustering is playing a more and more important role. From the perspective of the data centers, it's necessary to handle numerous users' requests every day, of which the computational jobs are of high concurrency, and the requests for computing resources are of great variance. To make the cloud computing more efficient, it's necessary to predict and classify the computational jobs. To do so, we first analyze the characteristics of computational jobs and obtain a better perception of the arrival of the computational jobs as well as the requests for computing resources with machine learning approaches. The results obtained here may provide some guidance or optimization for the scheduling of computing tasks and the resources.  It is noted that this work can also be extended to that of edge network computing or fog computing if the requests to fog computing center are too much~\cite{liu2017rf}\cite{chiang2016fog}. 
\subsection{Dataset}
In this paper we evaluate our proposed prediction method and do clustering on ClusterData2011\_2~\cite{reiss2011google} which is a dataset released by Google in 2011 and updated in 2014, containing the practical data of more than 10 thousands computing servers of Google's data centers in one month. The records of the machine events, job events, task events, resource usage, etc. are included.
\subsection{Related Works}
Some related works in \cite{chen2014trace} demonstrated that computational jobs and the resource requests have the characteristics of non-linearity, non-stationarity and self-similarity. It also put forward a fractal model to predict the arrival of the jobs and the resource requests, resulting in a better performance compared to auto-regression method. In~\cite{prevost2011prediction}, it employed a simple neural model, a 3-layer MLP (multilayer perceptron), to make prediction. As for the classification, it applied the k-means algorithm which classified the computational jobs into 5 categories~\cite{alam2016analysis}. In \cite{abdul2017google}, the authors employed a hierarchy clustering approach based on more features and presented some observations from the view of execution time. What's more,~\cite{rasheduzzaman2014task} applied the shape classification and workload characterization to the data and discussed the revelance between the execution time, resource usage and the classification results.

However, these traditional models sometimes cannot completely represent the characteristics of the data, especially at the spike points. To solve such kind of issues, we consider to apply some popular machine learning approaches such as LSTM neural networks which have strong representative ability and have been tested in considerable applications in prediction of time sequences.
\subsection{Contributions \& Structure of the Paper}
In this paper, we adopt a machine learning approach, LSTM neural network, to make the prediction of the computational jobs. As for the clustering, we select the features that can be gained before the execution so that we can have a priori comprehension of the computational jobs. Here BIRCH, Balanced Iterative Reducing and Clustering using Hierarchies,  is employed to classify the jobs. What's more, the revelance between the classification results and the execution data is also explored to see whether the classification using the priori features makes sense or not.

The main contributions of this work can be summarized as follows: (1) Two LSTM neural networks are built, while one is used to predict the arrival interval of the jobs and the other is used to predict the aggregated requests for computing resources within a certain time slot. It shows that the accuracy is much better compared to the existing similar works such as auto-regression and fractal modeling technique; (2) According to the features of the jobs that can be gained before execution, we make a unsupervised hierarchical clustering of jobs and evaluate it numerically using Davies-Bouldin indicator and Silhouettes indicators. (3) With the ahead-of-execution classification, we investigate the relation between the results and the practical executions.

The paper is organized as follows: In section \uppercase\expandafter{\romannumeral2}, we'll give some definitions of the data in cloud computing that we're interested in. Then in section \uppercase\expandafter{\romannumeral3}, we'll introduce the LSTM network in short and show our method as well as the results of the prediction. In section \uppercase\expandafter{\romannumeral4}, BIRCH algorithm is introduced first and some new discoveries from the clustering by BIRCH is presented. Finally, we'll conclude this work and give some further research directions in section \uppercase\expandafter{\romannumeral5}.

%% file: 3lstm.tex
\section{Some Definitions Related to the Data}

In this section, we first introduce some necessary definitions. The arrival interval time sequence of the computational jobs, $\left\{t_{interval}(i)\right\}$, is defined as:
\begin{equation}
t_{interval}(i)=
\begin{cases}
0 & i=1 \\
T_{arrival}(i)-T_{arrival}(i-1) & else
\end{cases}
\end{equation}
where $T_{arrival}(i)$ is the arrival time of the $i$-th computing job in cloud center. The resource requests of each job can be expressed in the following tensor form:
\begin{equation}
\boldsymbol r(i)=[r_{CPU}(i),r_{mem}(i),r_{disk}(i)]
\end{equation}
where $r_{CPU}(i),r_{mem}(i)$ and $r_{disk}(i)$ represent the $i$-th job's request for CPU, RAM and disk capacity, respectively. Accordingly, the $k$-th computing request in certain time slot $t_{slot}$ is defined as:
\begin{equation}
\boldsymbol R(k)=\sum_{i\in A_k}[r_{CPU}(i),r_{mem}(i),r_{disk}(i)]
\end{equation}
where $A_k=\left\{i:\lfloor\frac{T_{arrival}(i)}{t_{slot}}\rfloor=k\right\}$, denotes the set of jobs in the $k$-th time slot and $\lfloor\cdot\rfloor$ means the floor of the positive real number. Note that the time slot is set to be 5 minutes in the following real data discussion.

For the classification section, we generate the feature vectors of each computing job from the task events, containing the inter arrival time, the degree of parallelism and the request of computing resources. The feature vector of the $i$-th computing job is defined as :
\begin{equation}\label{feature}
\boldsymbol f(i)=[t_{interval}(i),p(i),r_{CPU}(i),r_{mem}(i),r_{disk}(i)]
\end{equation}
where $p(i)$ represents the degree of parallelism, i.e. the number of the subtasks of each computing job.

\section{LSTM-Based Prediction of computational jobs}
\subsection{A Brief Introduction of LSTM}
\begin{figure}[htbp]
\centering\includegraphics[width=0.48\textwidth]{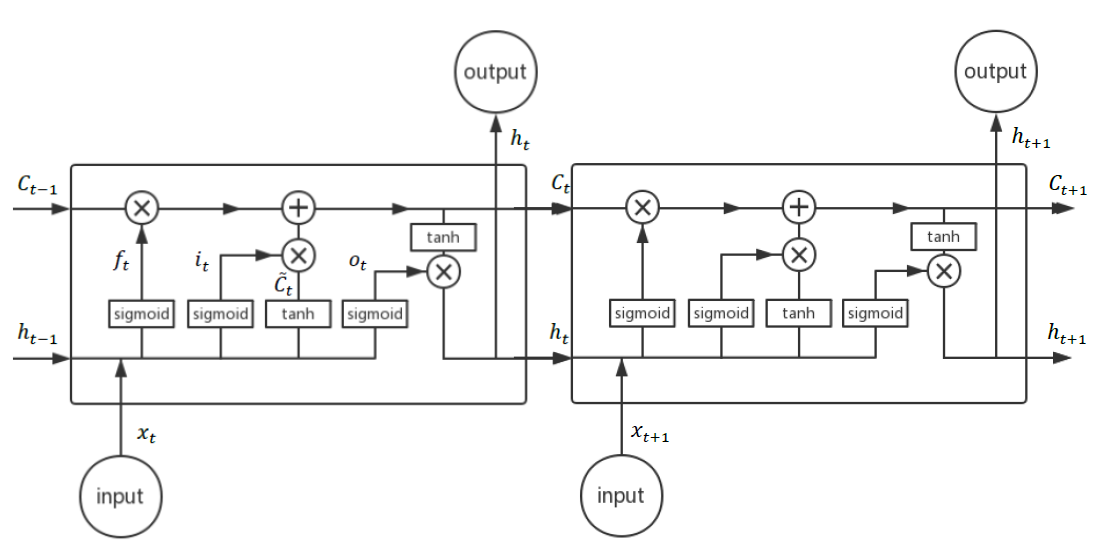}
\caption{The structure of LSTM cells}\label{lstm}
\end{figure}
It is well known that LSTM, long short term mermory network, proposed by Hochreiter and Schmidhuber in 1997,~\cite{hochreiter1997long} has become more popular in many fields of applications. Based on recurrent neural network (RNN), LSTM has the memory of time series by adding the forgetting gate. A typical structure of LSTM cells is shown in Fig.~\ref{lstm}. The progress of long short term memory can be described as: the output $h_{t}$ and the state $C_{t}$ of each cell contain the information of past and are transmitted between the adjacent LSTM cells. Through an end-to-end learning, we can obtain the parameters of the updating function $\tilde{C}_t$ as well as the gate functions $f_t$, $i_t$, and $o_t$. Also, all LSTM cells share the same parameters.

\subsection{The Design of Our Network}
\begin{figure}[htbp]
  \centering
  \subfigure[Predicting inter arrival time of computational jobs.]{
    \label{net:a} 
    \begin{minipage}[b]{0.24\textwidth}
      \centering
      \includegraphics[width=1\textwidth]{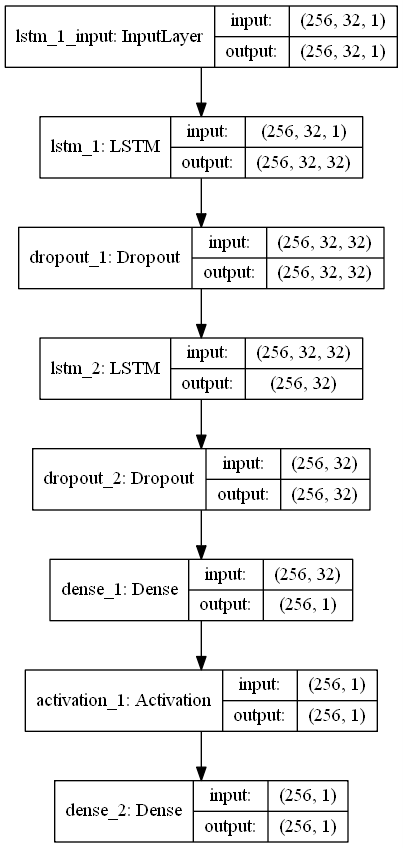}
    \end{minipage}}%
  \subfigure[Predicting aggregated requests of resources in each time slot.]{
    \label{net:b} 
    \begin{minipage}[b]{0.23\textwidth}
      \centering
      \includegraphics[width=1\textwidth]{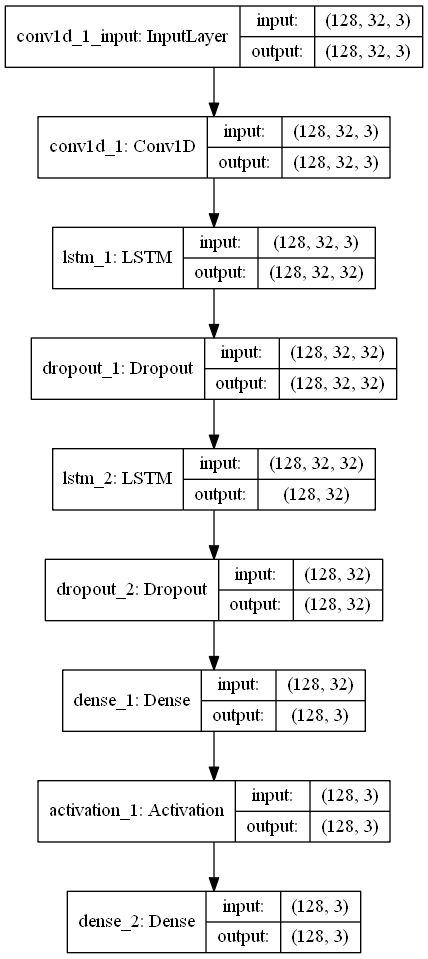}
    \end{minipage}}
  \caption{Network structure}
  \label{net} 
\end{figure}
To predict $t_{interval}$ and $\boldsymbol R(k)$ of computational jobs, we build two LSTM neural networks for each of them and the structures are shown in Fig.~\ref{net}. For the arrival interval time of computational jobs, it's a one-dimensional sequence, thus we adopt a 2-layer LSTM with the activation layer and full connected layer. While predicting the aggregated requests of resources in each time slot, we add a convolutional layer on the top to combine the information among the requests of CPU, RAM and disk capacity. Moreover, dropout~\cite{srivastava2014dropout} is add between each two layers to drop the connections randomly which simplifies the model so that the overfitting of the data~\cite{caruana2001overfitting} can be efficiently prevented.


\begin{figure*}[htbp]
  \centering
  \subfigure[Predicting the arrival interval.]{
    \label{res:sin} 
    \begin{minipage}[b]{0.49\textwidth}
      \includegraphics[width=1\textwidth]{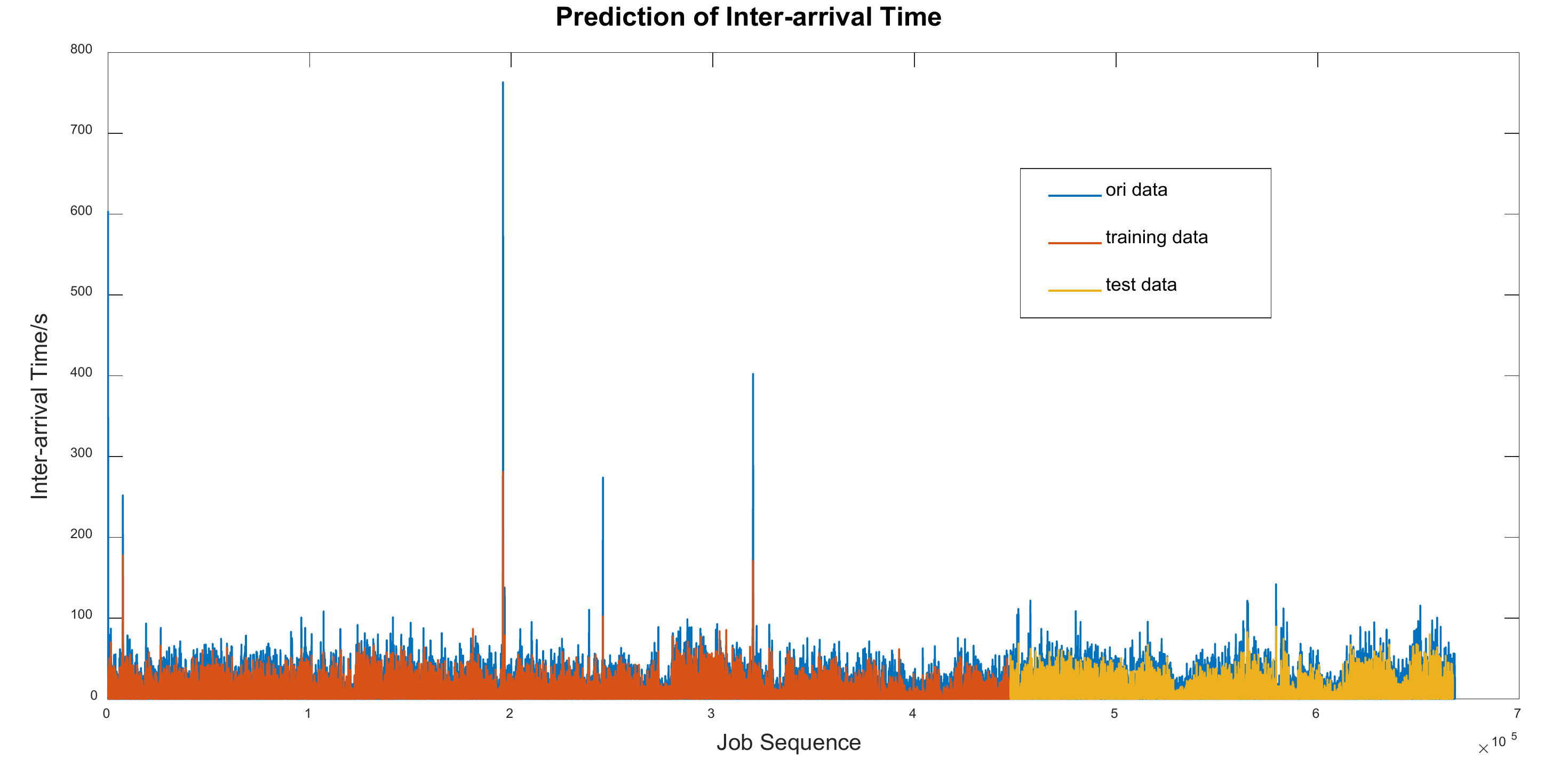}
    \end{minipage}}%
  \subfigure[Loss of predicting the aggregated requests.]{
    \label{res:loss} 
    \begin{minipage}[b]{0.49\textwidth}
      \includegraphics[width=1\textwidth]{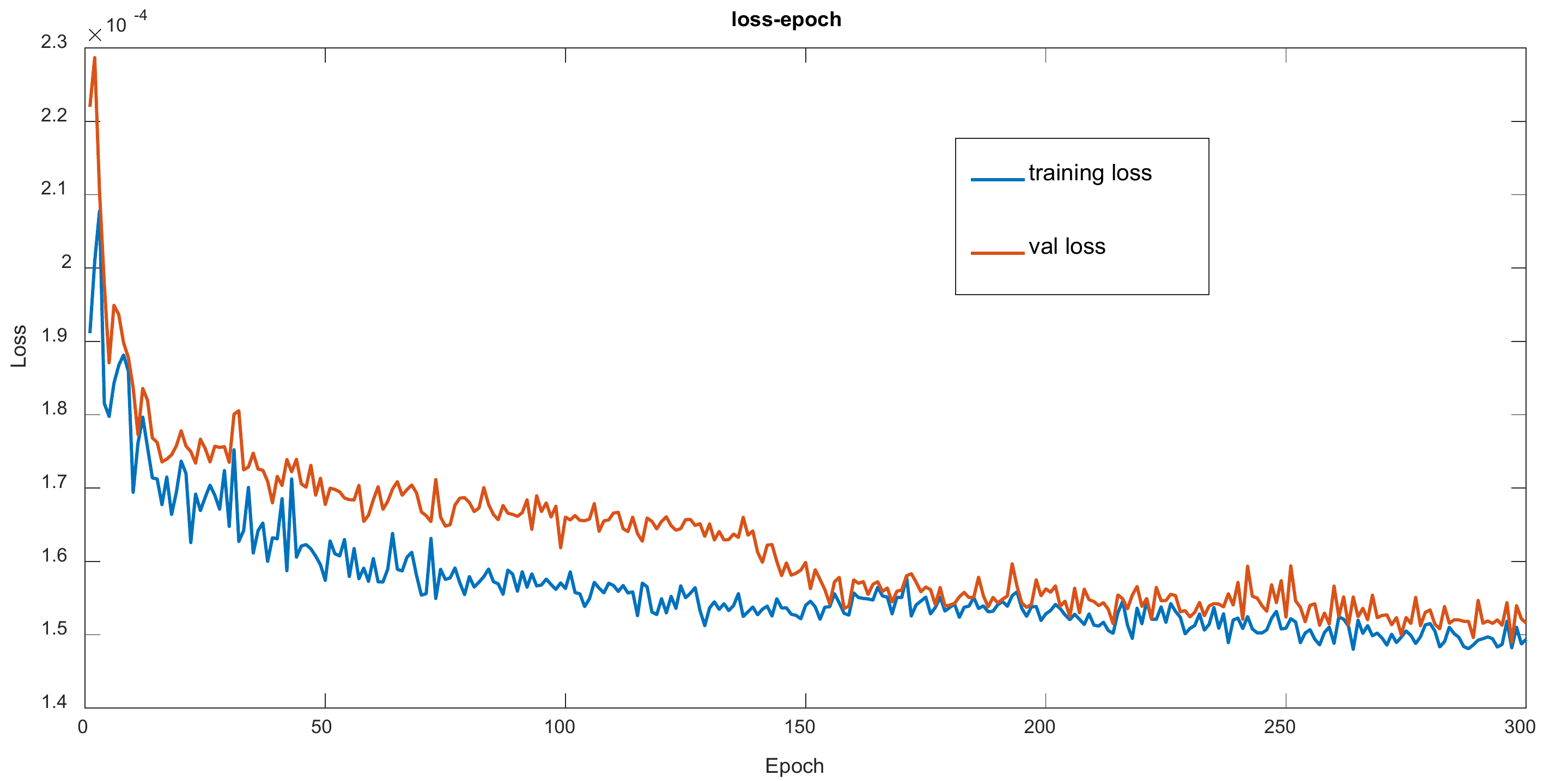}
    \end{minipage}}
  \subfigure[Predicting the aggregated computing requests.]{
    \label{res:mul} 
    \begin{minipage}[b]{0.49\textwidth}
      \includegraphics[width=1\textwidth]{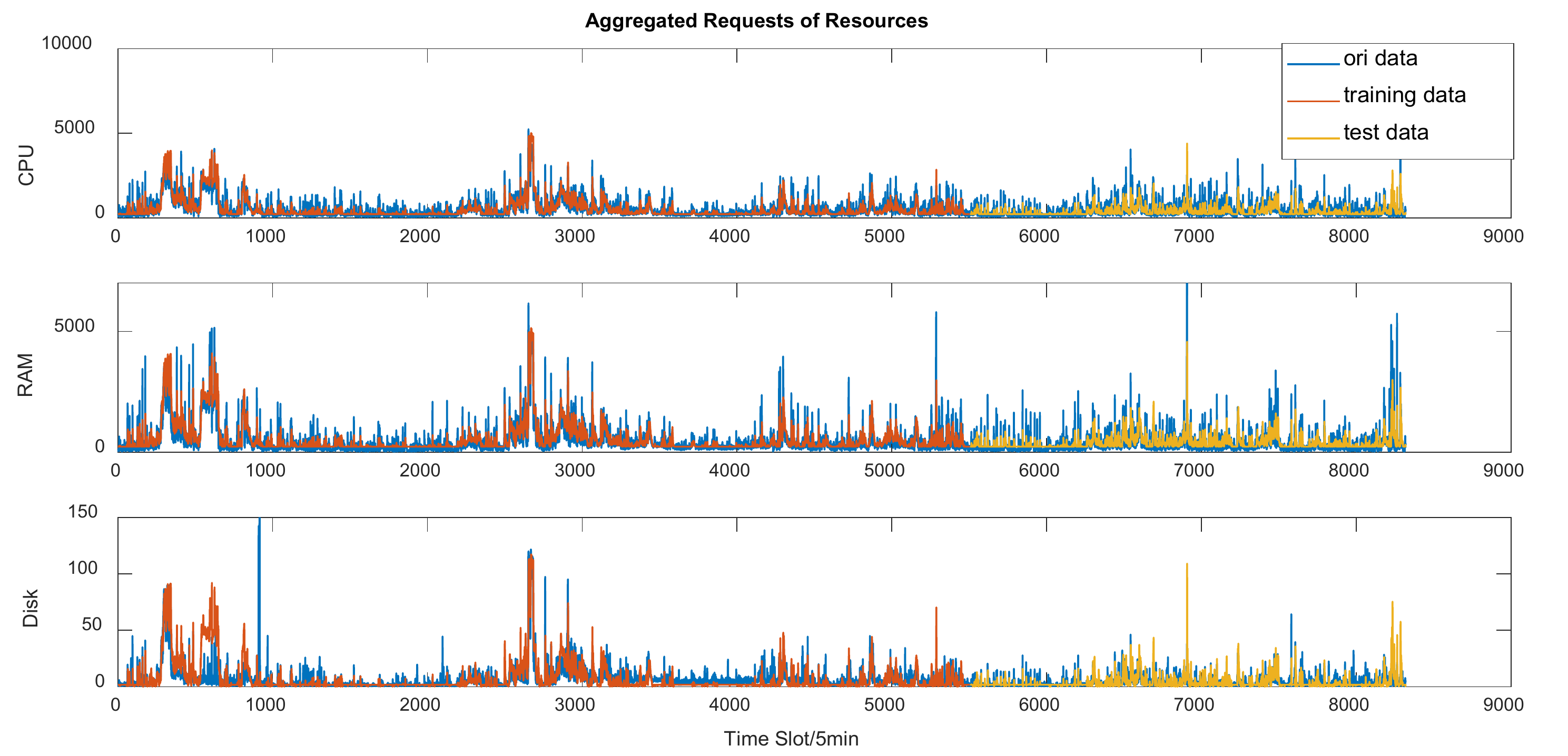}
    \end{minipage}}
  \subfigure[A compare with baselines.]{
    \label{res:cmp} 
    \begin{minipage}[b]{0.49\textwidth}
      \includegraphics[width=1\textwidth]{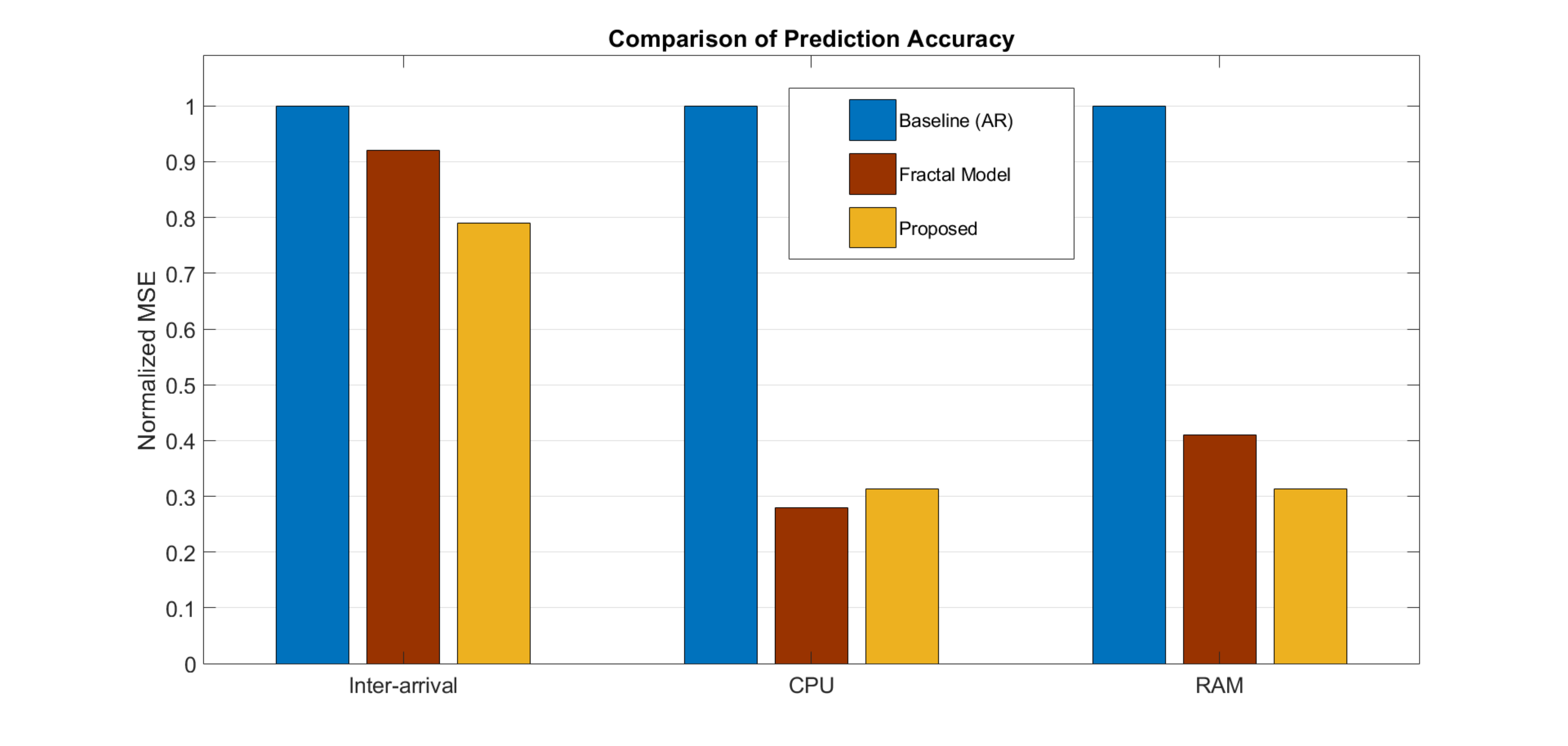}
    \end{minipage}}
  \caption{Performance of prediction.}
  \label{res} 
\end{figure*}
\subsection{The Evaluation of Prediction}
We implement the the LSTM-based neural networks proposed above in Python using Keras with Tensorflow as the back-end and evaluate the performance on Google Clusters dataset. The results are shown in Fig.~\ref{res}. In detail, Fig.~\ref{res:sin} is the result of the predicting the arrival interval time while the loss won't be displayed here because it's a single dimension sequence and the training converges after several epochs. The prediction of aggregated computing request is displayed in Fig.~\ref{res:mul} and most of the points especially the peak values are well predicted. In Fig.~\ref{res:loss} one can see that the prediction loss converges after 150 epochs and validation loss is close to training loss which means that there is no overfitting in our training. At last, we compare our method with related works on the dimensions that data centers are interested in, i.e. arrival interval, CPU and RAM capacity. Finally, we select AR method as the baseline and compare with the fractal method, overall we gain 21.6\%, 68.7\%, 68.5\% promotion in three dimensions, respectively compared to the baseline and the precisions are also 13.0\%, 9.6\% better than fractal model on arrival interval and the request of RAM, claimed in~\cite{chen2014trace}, while there is 3.3\% worse on the CPU prediction.

%% file: 4clustering.tex
\section{Clustering of computational jobs}
In this section, we will discuss on the clustering of the computational jobs. After removing the invalid data of task events, we select about $1/10$ computational jobs, 6.6k in total, as the samples and extract the 5 features of each jobs, i.e. the degree of parallelism, the arrival interval time and 3 computing requests. We generate the feature tensors as equation~(\ref{feature}) defined. It's a typical unsupervised problem thus we apply BIRCH algorithm to cluster the computational jobs.
\subsection{An Introduction of BIRCH}
\begin{figure}[htbp]
\centering\includegraphics[width=0.37\textwidth]{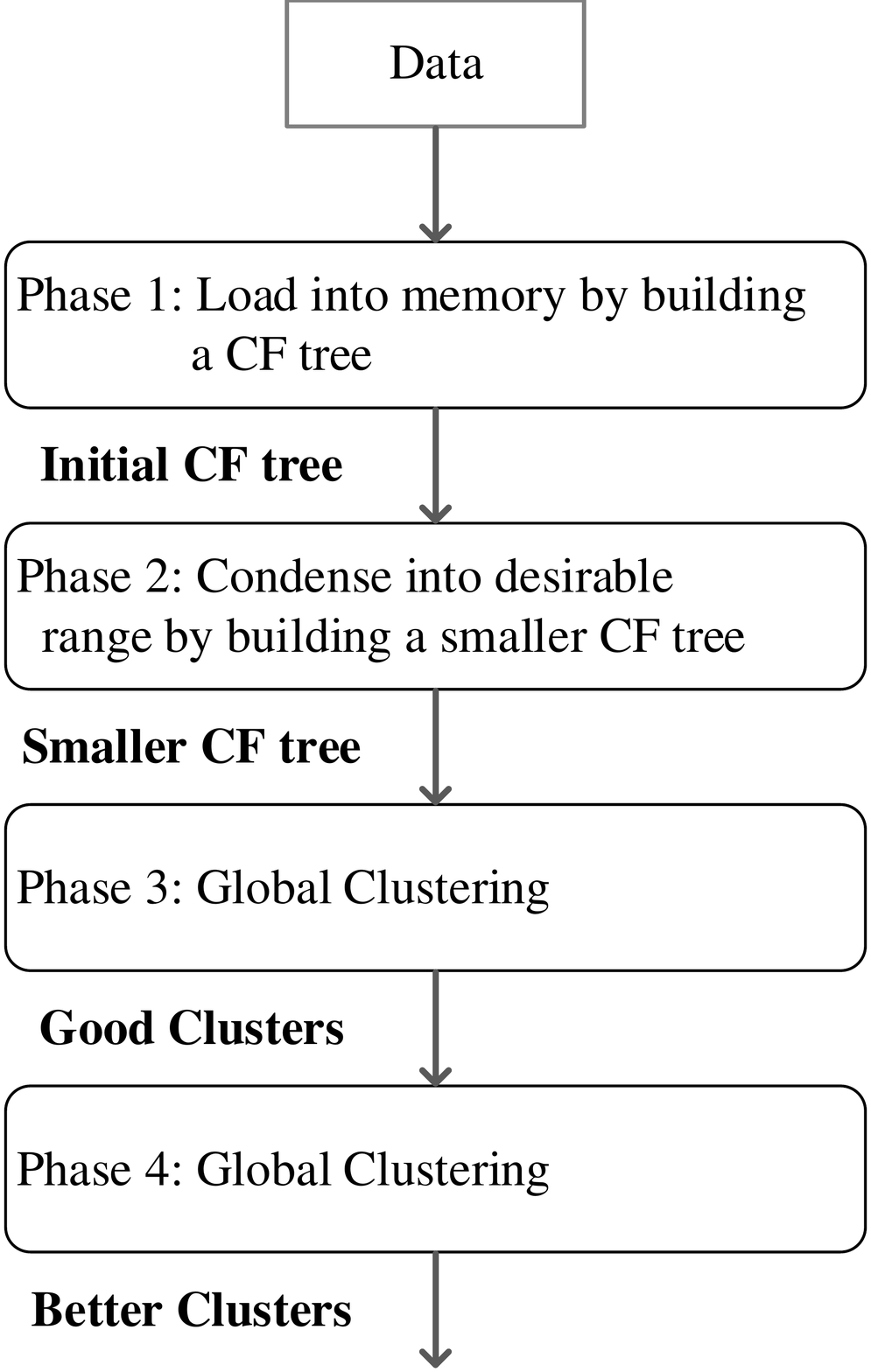}
\caption{The procedure of BIRCH}\label{birch}
\end{figure}
BIRCH (Balanced Iterative Reducing and Clustering using Hierarchies) algorithm, proposed by TIAN ZHANG et al.~\cite{zhang1996birch} generates a cluster feature tree (CF tree) of which each non-leaf node represents a sub-cluster and the samples are added in turn to update the profile of the CF tree. A brief procedure of BIRCH is shown in Fig.~\ref{birch}~\cite{zhang1997birch}. Instead of storing the samples in nodes, each node in CF tree only contains the principle of classification so that it fits our problem with enormous number of samples.
\subsection{Clustering Results and Evaluation}
\begin{figure}[htbp]
\centering\includegraphics[width=0.42\textwidth]{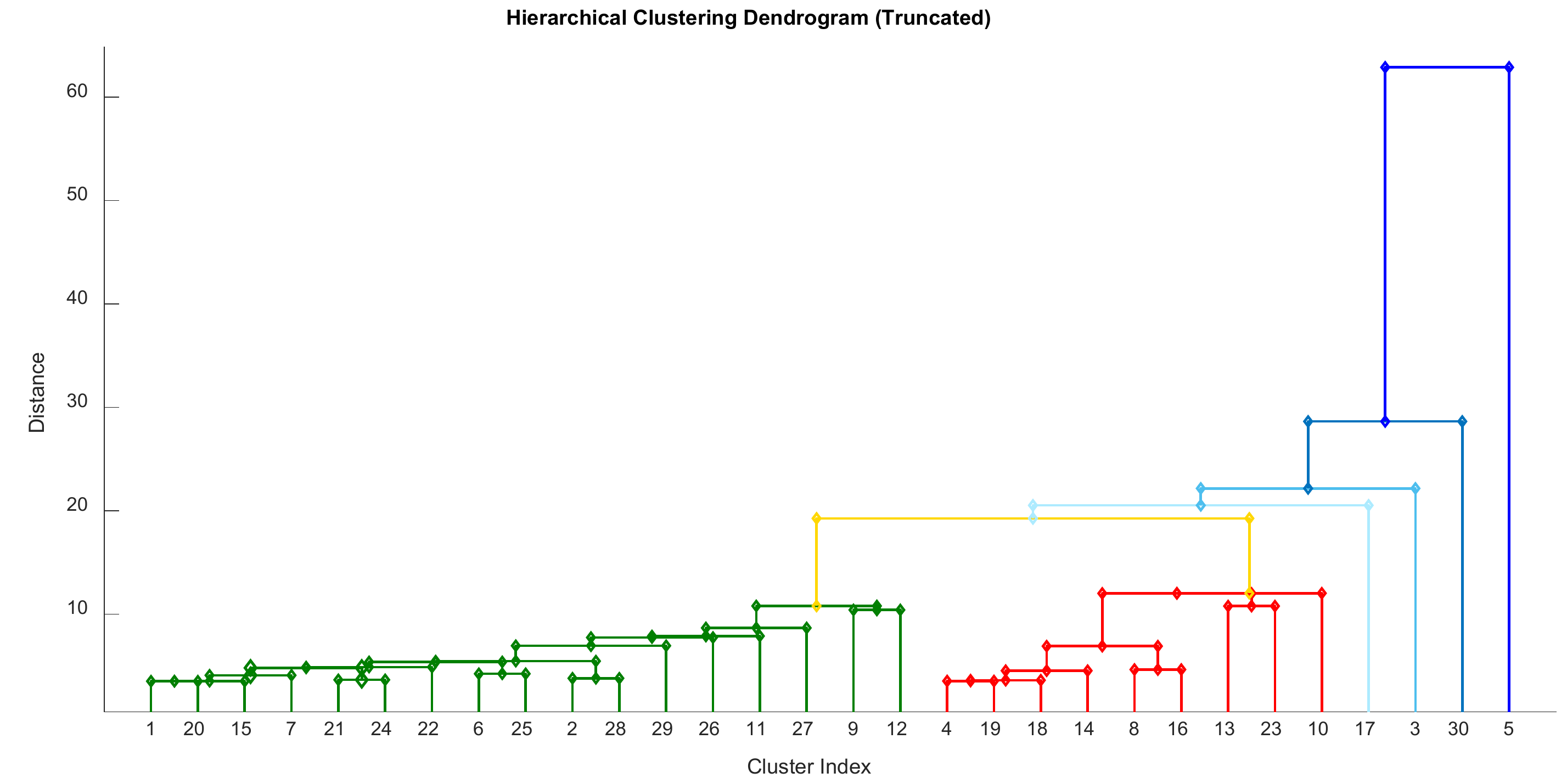}
\caption{A truncated result of BIRCH. The horizontal axis is the index of cluster and the vertical represents the distance between clusters.}\label{cluster}
\end{figure}
Fig.~\ref{cluster} gives a truncated clustering result of last 30 hierarchies and then we explain the result in both numeric and perception.
\subsubsection{Evaluation using Numeric Indicators}
\begin{figure}[htbp]
\centering
  \subfigure[Davies-Bouldin indicator of the last 30 hierarchies.]{
    \label{eva:dbi} 
    \begin{minipage}[b]{0.48\textwidth}
      \centering
      \includegraphics[width=1\textwidth]{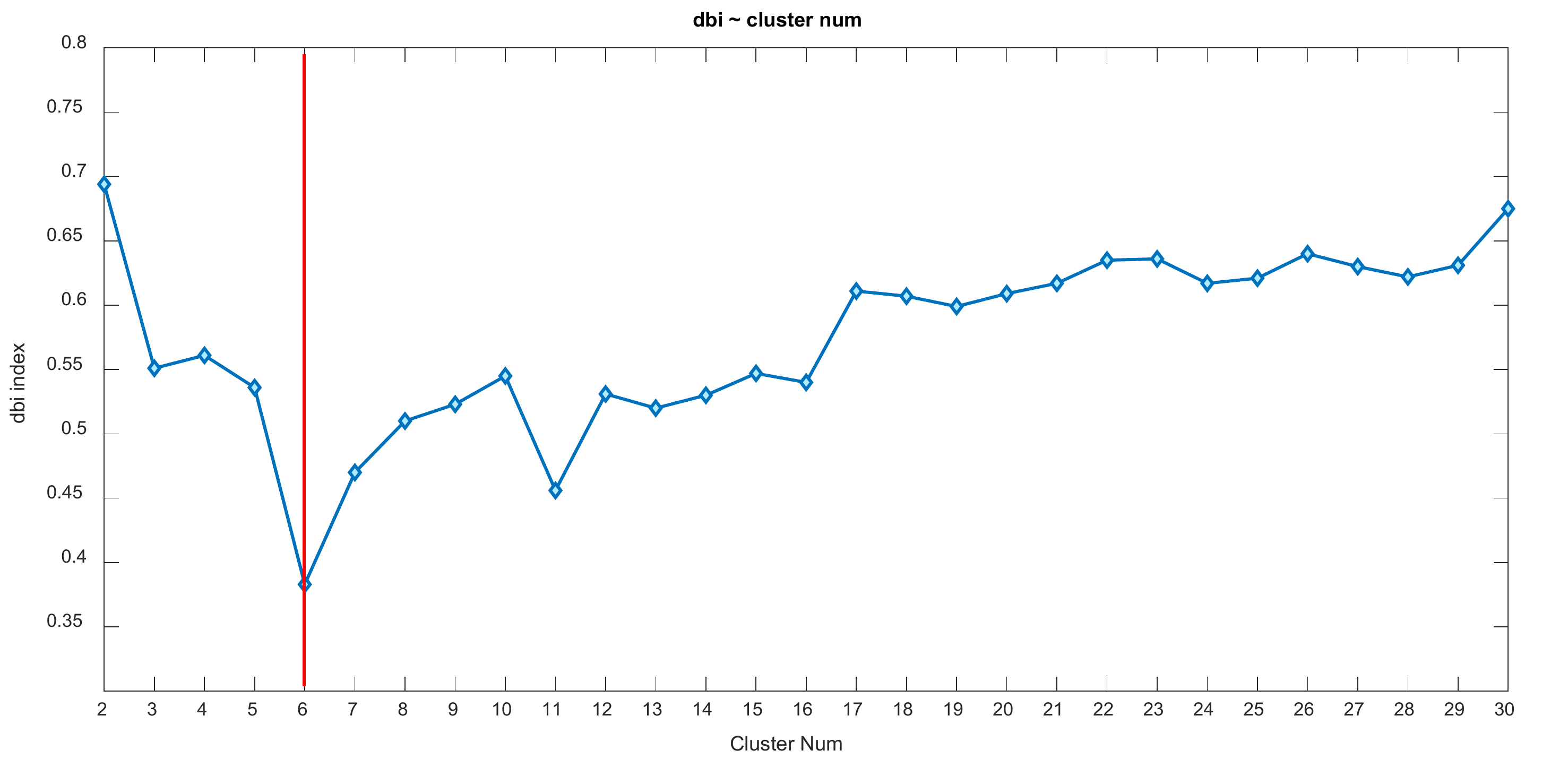}
    \end{minipage}}\\%
  \subfigure[Silhouettes indicator. Each horizontal line represents a sample.]{
    \label{eva:sil} 
    \begin{minipage}[b]{0.48\textwidth}
      \centering
      \includegraphics[width=1\textwidth]{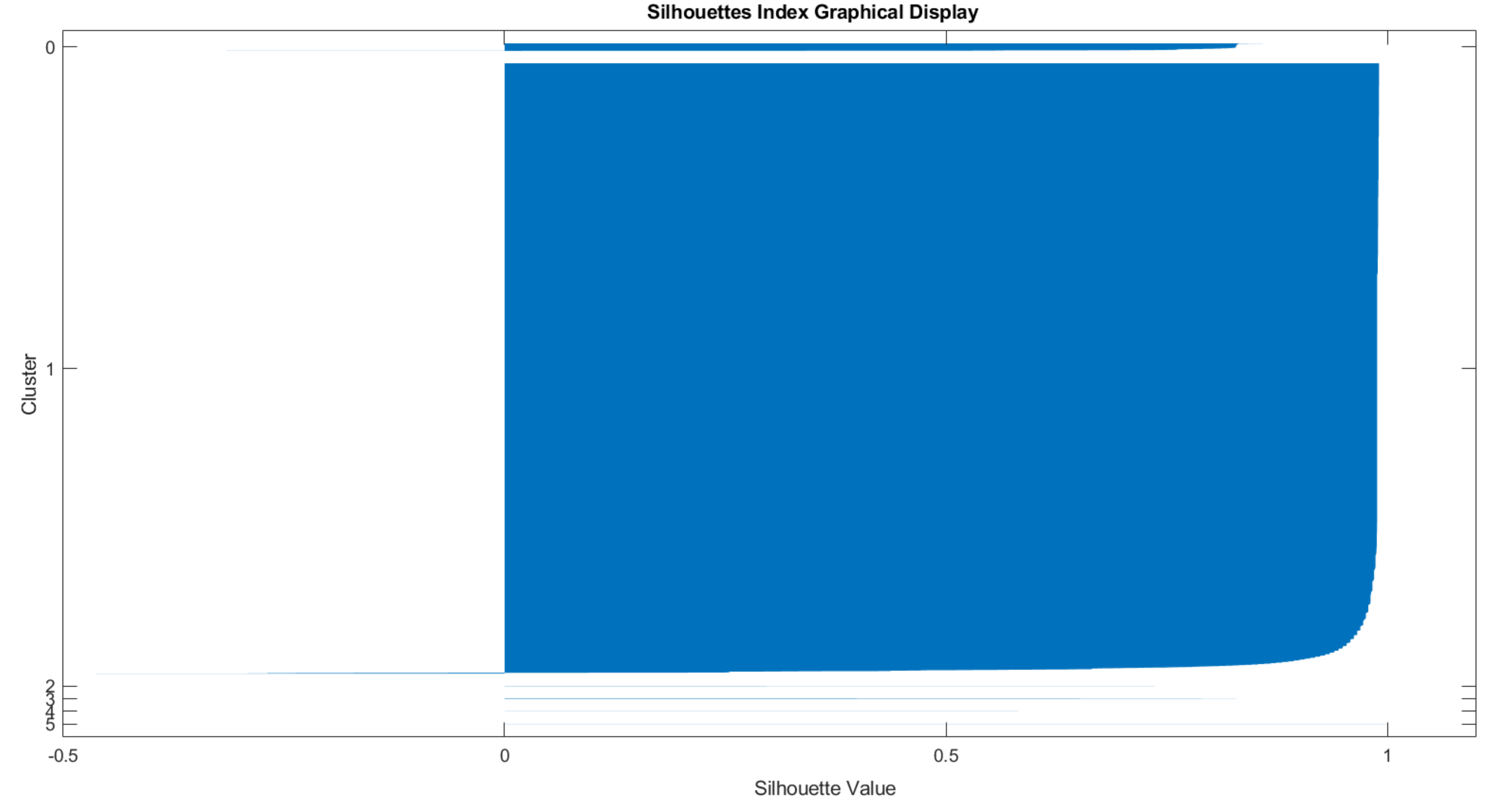}
    \end{minipage}}
  \caption{Numeric evaluatation of the clustering.}
  \label{eva} 
\end{figure}
To judge the effect of the BIRCH clustering the computational jobs, we first calculate the Davies-Bouldin indicator which is defined as~\cite{davies1979cluster}:
\begin{equation}
dbi=\frac{1}{N}\sum_{i=1}^{N}\max\left(\frac{\bar{C_i}+\bar{C_j}}{\|w_i-w_j\|}\right)
\end{equation}
where $N$ is the number of clusters while $\bar{C_i}$ denotes the average distance of the samples in $i$-th cluster and $w_i$ is the weight center of the $i$-th cluster. Davies-Bouldin indicator reflects the ratio of the internal distance to the external distance thus the smaller $dbi$ means the better effect of the clustering. Fig.~\ref{eva:dbi} shows the $dbi$ evolution in the last 30 hierarchies and therefore, we select 6 as the number of the clusters.
Additionaly, we evaluate the result of 6 clusters using Silhouettes indicator proposed by Peter J.Rousseeuw which is a visible way to see the effect of the clustering result. The indicator considers the non-similarity of two samples $O_i,O_j$, which has the definition as follows~\cite{rousseeuw1987silhouettes}:
\begin{equation}
d(O_i,O_j)=\sum_{k=1}^{m}\mathbbm{1}_{x_{ik}\neq x_{jk}}
\end{equation}
where $m$ is the dimension of each features and $\mathbbm{1}_{\cdot}$ is the indicator function. Then the Silhouettes indicator is defined as:
\begin{equation}
S(i)=\frac{b(i)-a(i)}{max\{a(i),b(i)\}}
\end{equation}
where $a(i)$ denotes the average non-similarity between sample $i$ and samples in the same cluster $A$ while $b(i)$ is the minimum average non-similarity between sample $i$ and samples in other clusters. Silhouettes indicator ranges from -1 to 1 where 1 corresponds to the the clustering with the best effect, and -1 means the worst. As shown in Fig.~\ref{eva:sil}, most of the samples have the $S(i)$ close to 1, and even few outliers have the $S(i)$ larger than -0.4. Thus, it indicates that the clustering is numerically effective according to these indicators.
\subsubsection{Evaluation in Interpretability}
\begin{figure*}[htbp]
\centering
  \subfigure[Arrival interval vs CPU vs RAM.]{
    \label{clu:bad} 
    \begin{minipage}[b]{0.49\textwidth}
      \centering
      \includegraphics[width=1\textwidth]{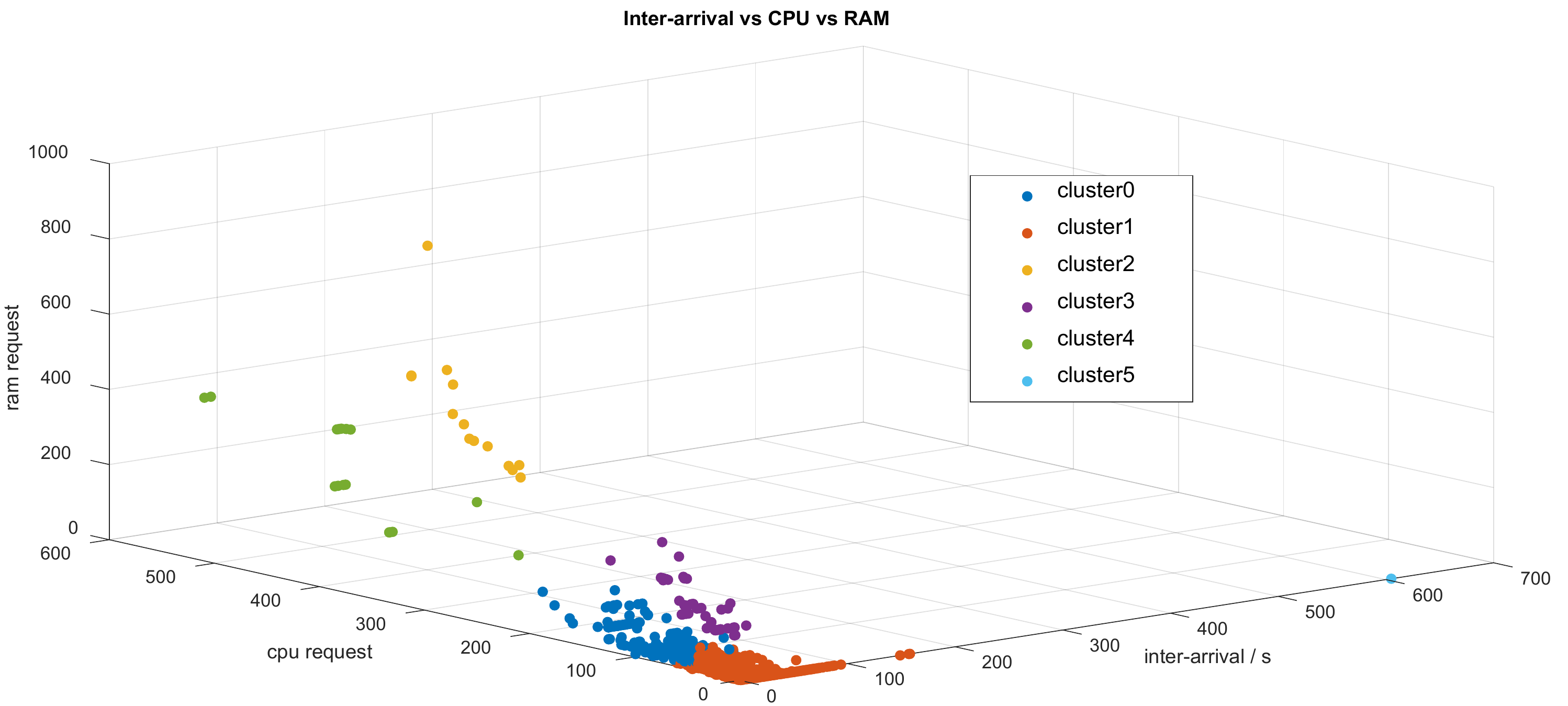}
    \end{minipage}}%
  \subfigure[CPU vs RAM vs disk.]{
    \label{clu:good} 
    \begin{minipage}[b]{0.49\textwidth}
      \centering
      \includegraphics[width=1\textwidth]{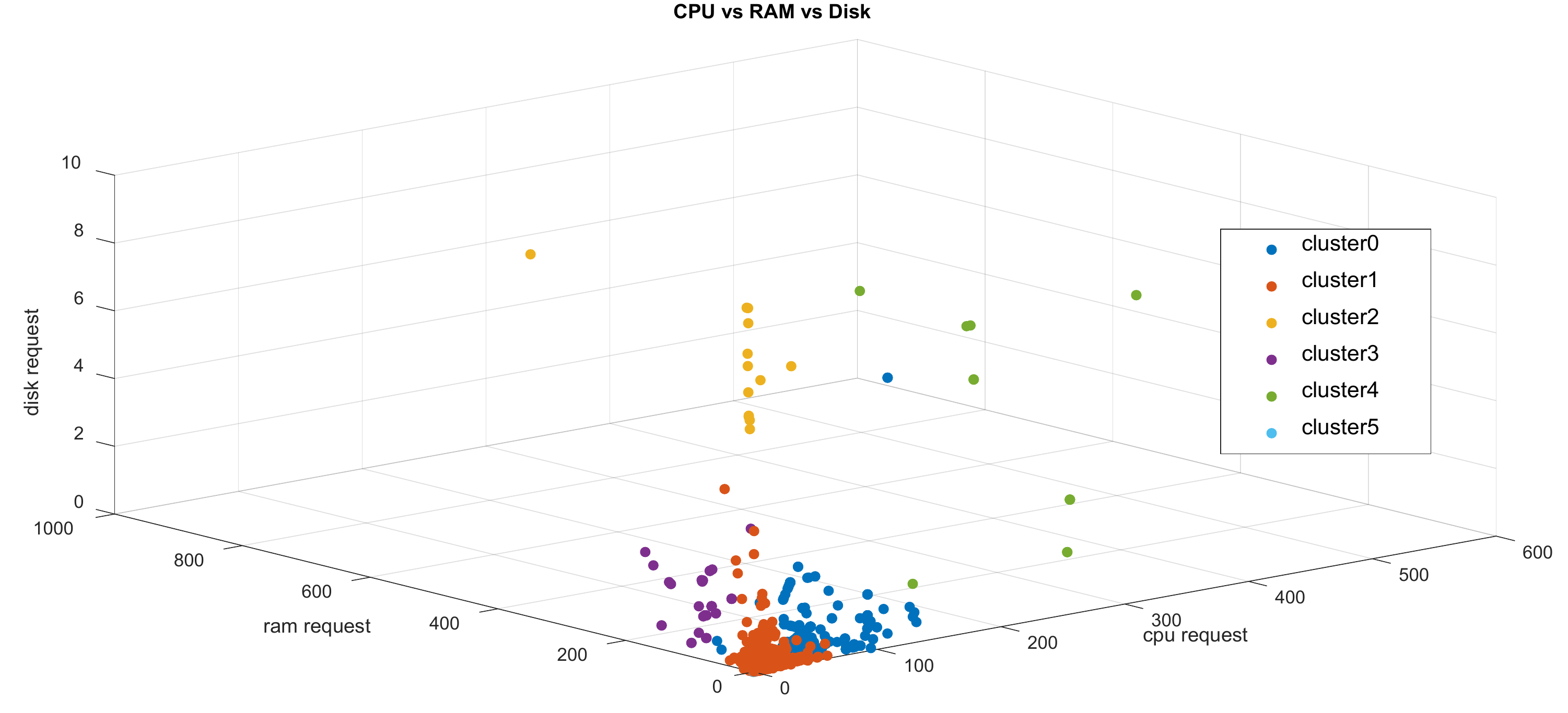}
    \end{minipage}}
  \caption{Plots on different dimensions of features.}
  \label{clu} 
\end{figure*}
\begin{table*}[htbp]
\centering
\begin{tabular}{ccc}
\hline
Job Cluster& Count& Features\\
\hline
Cluster0& 821& medium CPU request, little RAM and disk capacity\\
Cluster1& 65711& little CPU and RAM request, partly medium disk capacity\\
Cluster2& 14& large CPU request, RAM, and disk capacity\\
Cluster3& 49& little CPU request and disk capacity, medium RAM capacity\\
Cluster4& 16& large CPU request, medium RAM capacity, no obvious feature on disk capacity\\
\hline
\end{tabular}
\caption{A brief description of the clustering results.}
\label{tab}
\end{table*}
In this phase, we plot the distribution of several features in one figure to see whether the result corresponds to the features of each clusters. As shown in Fig.~\ref{clu:bad}, different clusters can be clearly distinguished on CPU and RAM except the arrival interval time which is similar when parallelism is considered. For this reason, it makes sense that the clustering result on the arrival interval time and the parallelism cannot be interpreted explicitly. Besides, the cyan point denoting Cluster 5 in the figure is a single point with extremely large arrival interval and seems no difference on other dimensions so Cluster 5 is removed in the further discussion. We display CPU-RAM-disk distribution in Fig.~\ref{clu:good} and samples of same cluster assemble in separate area on each dimension so different clusters can be well distinguished in these three dimensions and the characteristics of 5 clusters can be summarized in Tab.~\ref{tab}.
\subsubsection{Relevance with Practical Execution Data}
\begin{figure}[htbp]
\centering\includegraphics[width=0.48\textwidth]{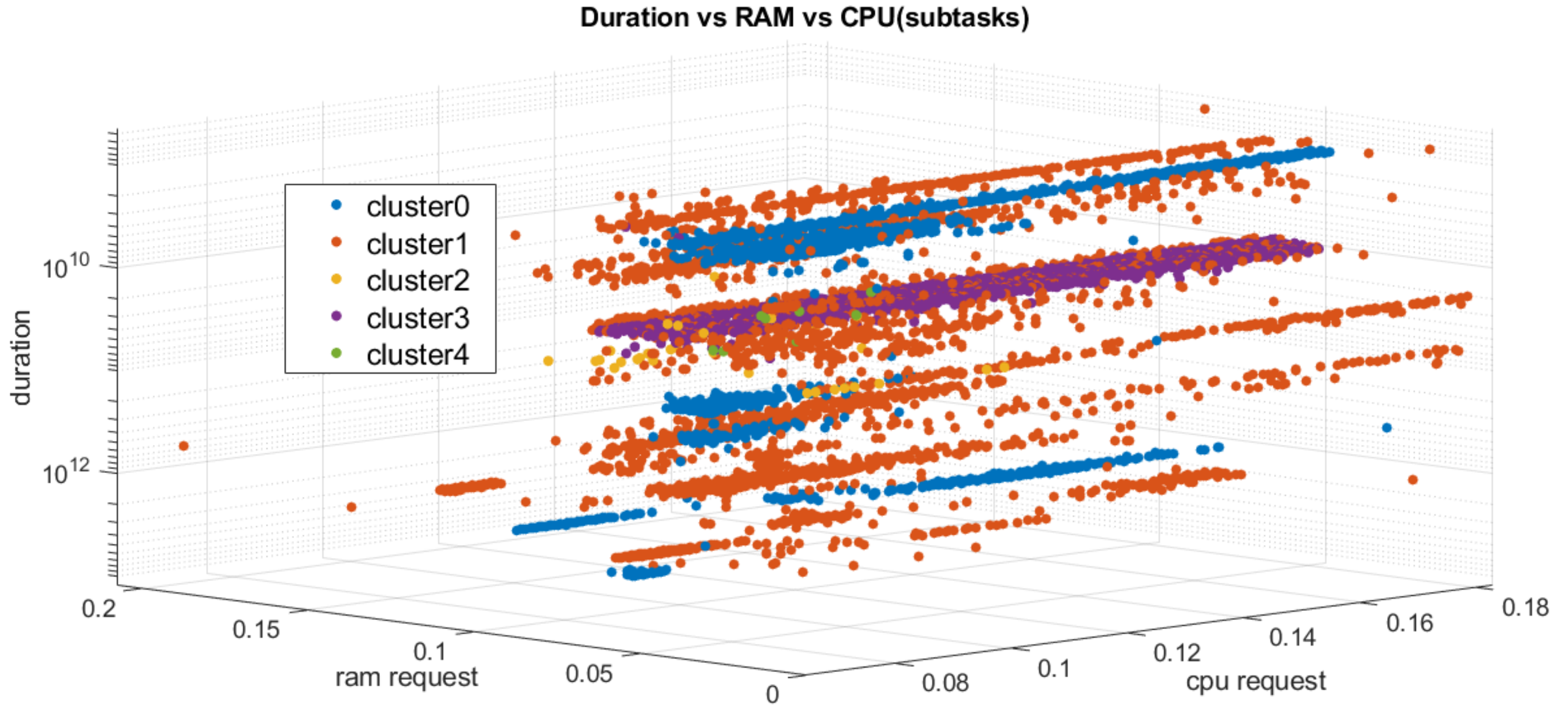}
\caption{Features of practical execution data.}\label{cluana}
\end{figure}
Moreover, it should be emphasized that the features we used in clustering are extracted before the executions which means we can classify the jobs basically ahead of executions. So that it becomes more clearly to observe the relevance between the clustering results and the practical execution, and fin out whether our classification has some insights in practical cases. We notice that each computing job will be divided into a number of subtasks while executed. Then we extract the execution data of subtasks and similarly, we plot the same features of the subtasks in Fig.~\ref{cluana}. Only a part of the computing subtasks are shown and we can see that subtasks of different clusters also have clear borders which proves that the clustering using ahead-of-execution features can reflect the practical executions. It indicates that with the ahead-of-execution clustering, the subtasks divided from computational jobs can also be classified, which may improve the computing efficiency greatly.

%% file: 5conclusion.tex
\section{Conclusion and Future Work}
In this paper, we implemented an LSTM-based method to predict the arrival interval as well as the aggregated computing requests of jobs in cloud computing cases. The evaluation on Google Clusters shows that we get accuracy improvement on the prediction of several factors comparing to some known methods. We also did a hierarchy clustering of the computational jobs with BIRCH algorithm and proved that the results make sense both numerically and practically. In addition, we found that using the features extracted ahead of executions, the results of classification can exactly reflect the execution of these jobs.

This work on the prediction and classification of computational jobs provides insights of the traits of cloud computing which may stimulate us to explore more machine learning approaches to solve the simulation, task scheduling, load balancing and other optimization problem in cloud computing.